\documentclass[runningheads]{llncs}
\usepackage{graphicx}
\usepackage{amsmath}
\usepackage{amssymb}
\usepackage{booktabs}
\usepackage{multirow}
\usepackage{pifont}
\usepackage{nicefrac}
\usepackage{comment}
\usepackage{wrapfig}
\usepackage{subfig}
\usepackage[pagebackref,breaklinks,colorlinks]{hyperref}

\mathchardef\mhyphen="2D

\usepackage{color}
\usepackage{xcolor}
\usepackage{caption}
\usepackage{mmstyle}
\usepackage[capitalize]{cleveref}
\crefname{section}{Sec.}{Secs.}
\Crefname{section}{Section}{Sections}
\Crefname{table}{Table}{Tables}
\crefname{table}{Tab.}{Tabs.}

\newcommand{\tablestyle}[2]{\setlength{\tabcolsep}{#1}\renewcommand{\arraystretch}{#2}\centering\footnotesize}
\newlength\savewidth\newcommand\shline{\noalign{\global\savewidth\arrayrulewidth
		\global\arrayrulewidth .8pt}\hline\noalign{\global\arrayrulewidth\savewidth}}

\usepackage[accsupp]{axessibility}  % Improves PDF readability for those with disabilities.

\begin{document}
\pagestyle{headings}
\mainmatter

\title{DG-STGCN: Dynamic Spatial-Temporal Modeling for Skeleton-based Action Recognition}

\titlerunning{DG-STGCN} 
\authorrunning{Haodong Duan, Jiaqi Wang, Kai Chen, Dahua Lin} 

\author{Haodong Duan\inst{1, 2} \and
Jiaqi Wang\inst{2} \and Kai Chen\inst{2} \and Dahua Lin\inst{1, 2}}

\institute{The Chinese University of Hong Kong \and Shanghai AI Lab}

\maketitle

\begin{abstract}

Graph convolution networks (GCN) have been widely used in skeleton-based action recognition. 
We note that existing GCN-based approaches primarily rely on prescribed graphical structures
(\emph{i.e.}, a manually defined topology of skeleton joints), 
which limits their flexibility to capture complicated correlations between joints. 
To move beyond this limitation, we propose a new framework for skeleton-based action recognition, 
namely Dynamic Group Spatio-Temporal GCN (DG-STGCN). 
It consists of two modules, DG-GCN and DG-TCN, respectively, for spatial and temporal modeling. 
In particular, DG-GCN uses learned affinity matrices to capture dynamic graphical structures instead of relying on a prescribed one, 
while DG-TCN performs group-wise temporal convolutions with varying receptive fields and 
incorporates a dynamic joint-skeleton fusion module for adaptive multi-level temporal modeling. 
On a wide range of benchmarks, including NTURGB+D, Kinetics-Skeleton, BABEL, and Toyota SmartHome, 
DG-STGCN consistently outperforms state-of-the-art methods, often by a notable margin.

\end{abstract}
% Use present tense by default

\section{Introduction}

Human action recognition is a central task in video understanding.
For videos, various modalities derived from the rich multimedia are beneficial to the recognition task, 
including RGB~\cite{carreira2017quo,duan2020omni,feichtenhofer2020x3d,feichtenhofer2019slowfast,tran2019video,tran2018closer}, 
optical flow~\cite{piergiovanni2019representation,sevilla2018integration,simonyan2014two}, 
and human skeletons~\cite{choutas2018potion,duan2021revisiting,liu2020disentangling,shi2019two,yan2018spatial}. 
Among them, skeleton-based action recognition has attracted increasing attention
due to its action-focusing nature and robustness against complicated backgrounds or different lighting conditions.
In contrast to other modalities, 
skeleton data are highly compact and abstract, containing only 2D~\cite{carreira2017quo,jhuangICCV2013} or
3D~\cite{Das_2019_ICCV,liu2020ntu,punnakkal2021babel,shahroudy2016ntu} human joint coordinates. 
For action recognition, the compactness of skeleton data leads to robust features and efficient processing.

Since Yan \emph{et al.}~\cite{yan2018spatial} first proposes ST-GCN to model skeleton motion patterns with a spatial-temporal graph, 
GCN-based approaches have quickly become the most popular paradigm for skeleton-based action recognition.
In ST-GCN, spatial modeling and temporal modeling are performed separately by spatial graph convolutions and temporal convolutions. 
Spatial graph convolutions fuse features of different joints according to a manually defined topology, 
presented as a sparse adjacency matrix indicating the direct connections between skeleton joints.
For temporal modeling, temporal 1D convolutions are applied to each joint in parallel to model the joint-specific motion patterns.

\begin{figure}[t]
    \centering
    \captionsetup{font=small}
	\includegraphics[width=\linewidth]{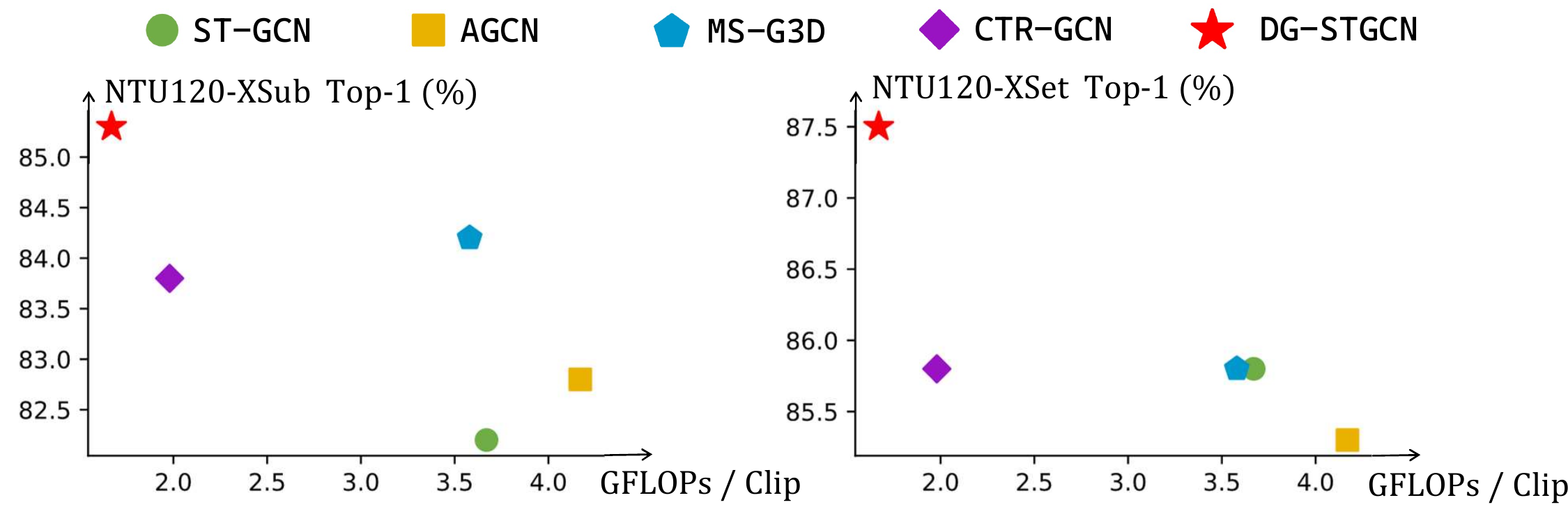} 
	\vspace{-6mm}
	\caption{\textbf{GFLOPs \emph{vs.} accuracies on two NTURGB+D 120 benchmarks\protect\footnotemark. } }
	\label{fig-teaser}
    \vspace{-6mm}
\end{figure}
\footnotetext{We report the single model accuracy trained on the joint modality. Details in Table~\ref{tab-aug}. }

Following works~\cite{chen2021channel,shi2019two,ye2020dynamic} propose to improve the spatial modeling 
with a learnable topology refinement. 
Though demonstrating good effectiveness, these approaches still heavily rely on the manually defined skeleton topology, 
requiring painstaking customizations for different datasets.
More importantly, due to the intrinsic cooperation among joints when performing actions, 
fully learnable coefficient matrices, rather than a prescribed graphical structure, are more suitable to model complicated joint correlations. 
Meanwhile, previous GCN-based approaches mostly utilize temporal convolutions with a fixed receptive field to model the joint-level motions in a certain temporal range, 
overlooking the benefits of modeling multi-level (\emph{i.e.} joint-level \texttt{+} skeleton-level) motion patterns within a dynamic temporal receptive field.

Revisiting the limitations of current works, we propose a novel GCN architecture for skeleton-based action recognition, namely DG-STGCN.
It enables group-wise dynamic spatial-temporal modeling for skeleton data.
In DG-STGCN, spatial modeling and temporal modeling are performed separately by dynamic group GCNs (\textbf{DG-GCNs}) 
and dynamic group temporal ConvNets (\textbf{DG-TCNs}), respectively.
Our proposed framework has three appealing properties. 
\textbf{First}, DG-STGCN is purely based on learnable coefficient matrices for spatial modeling, 
eliminating the cumbersome procedure to define a good joint topology manually. 
\textbf{Second}, the dynamic group-wise design enables the dynamic spatial-temporal modeling of the skeleton motion 
with diversified groups of graph convolutions and temporal convolutions, improving the representation capability and flexibility. 
\textbf{Third}, DG-STGCN achieves significant improvements on multiple benchmarks while preserving the model efficiency.

In DG-STGCN, both DG modules first transform the input skeleton features to $N$ feature groups (channel width reduced to $1/N$), 
and then perform spatial or temporal modeling independently for each feature group. 
In DG-GCN, each feature group has its own dynamic coefficient matrix\footnote{
    $N=8$ in experiments. For a manually defined topology, $N$ can only be 1 or 3.} for inter-joint spatial modeling.
Each coefficient matrix is a dynamic summation of three data-driven components, 
including one shared matrix that models the joint correlations across all samples (the static component) 
and two sample-specific matrices (the dynamic components) with different designs.
In experiments, DG-GCN demonstrates great capability, surpassing multiple variants of graph convolutions in both efficacy and efficiency. 
For temporal modeling, we propose a dynamic group temporal ConvNet (DG-TCN) with diversified receptive fields,
which further adopts a dynamic joint-skeleton fusion (D-JSF) module to fuse both joint-level and skeleton-level motion patterns (D-JSF) in various temporal ranges. 
Specifically, it models joint-level and skeleton-level features parallelly with a multi-group temporal ConvNet, 
where each group extracts motion patterns in a different temporal receptive field. 
The joint-level and skeleton-level features are dynamically fused into the joint-skeleton motion features with learnable joint-specific coefficients.
With computational efficiency preserved, DG-TCN is of great temporal modeling capability. 
Meanwhile, the proposed D-JSF module masters multi-level temporal modeling with negligible additional cost. 

With improved flexibility comes increased overfitting risk. 
We further propose to adopt Uniform Sampling as a strong temporal data augmentation strategy, 
which randomly samples a subsequence from the skeleton sequence data to generate highly diversified training samples. 
The strategy leads to consistent improvements across multiple backbones and benchmarks
and is especially beneficial for DG-STGCN. 

Extensive experiment results highlight the effectiveness of the dynamic group design and the good practices we proposed. 
Our DG-STGCN outperforms all previous state-of-the-art methods across multiple skeleton-based action recognition benchmarks, 
including NTURGB+D~\cite{liu2020ntu,shahroudy2016ntu}, Kinetics-Skeleton~\cite{carreira2017quo,yan2018spatial}, BABEL~\cite{punnakkal2021babel}, and Toyota SmartHome~\cite{Das_2019_ICCV}. 
\section{Related Works}

\subsection{Graph Neural Networks }
To process non-Euclidean structured data (like arbitrarily structured graphs), 
Graph Neural Networks \cite{bruna2013spectral,defferrard2016convolutional,hamilton2017inductive,hammond2011wavelets,kipf2016semi,velivckovic2017graph,xu2018powerful} (GNNs) are widely adopted and extensively explored. 
GNNs can be generally categorized as spectral GNNs and spatial GNNs.
Spectral GNNs~\cite{bruna2013spectral,hammond2011wavelets,henaff2015deep} apply convolutions on the specrtal domain.
They assume fixed adjacency among all samples, limiting the generalizability to unseen graph structures.
Spatial GNNs~\cite{duvenaud2015convolutional,hamilton2017inductive,velivckovic2017graph}, in contrast, 
perform layer-wise feature updates for each node by local feature fusion and activation. 
Most GCN approaches for skeleton-based action recognition follow the spirit of Spatial GNNs.
They construct a spatial-temporal graph based on the skeleton data, 
apply convolutions for feature aggregation in a neighborhood, and perform layer-wise feature updates.

\subsection{Skeleton-based Action Recognition }
Human skeletons are robust against backgrounds or illumination changes and 
can also be obtained with sensors~\cite{zhang2012microsoft} or pose estimators~\cite{8765346,rogez2019lcr,sun2019deep}.
Various paradigms have been designed for skeleton-based action recognition. 
Early approaches \cite{hussein2013human,vemulapalli2014human,wang2012mining} design hand-crafted features 
to capture joint motion patterns and feed them directly to downstream classifiers. 
With the prosperity of deep learning, 
the following methods consider skeleton data as time series and process them using 
Recurrent Neural Networks~\cite{liu2016spatio,shahroudy2016ntu,song2017end,zhang2017geometric,zhu2016co} and 
Temporal Convolution Networks~\cite{ke2017new,kim2017interpretable}.
However, these approaches do not model the joint relationships explicitly, leading to inferior recognition performance. 
To mitigate this, GCN-based approaches~\cite{chen2021channel,liu2020disentangling,shi2019two,yan2018spatial} construct spatial-temporal graphs, which separately perform spatial modeling and temporal modeling.

An early application of GCN on skeleton-based action recognition is ST-GCN~\cite{yan2018spatial}.  
ST-GCN uses stacked GCN blocks to process skeleton data, while each block consists of a spatial module and a temporal module. 
Being a simple baseline, ST-GCN's instantiations of the spatial module and the temporal module are straightforward: 
the spatial module adopts sparse coefficient matrices derived from a pre-defined adjacency matrix for spatial feature fusion;
while the temporal module uses a single 1D convolution for temporal modeling. 
Following works inherit the main framework of ST-GCN and develop various incremental improvements on the design of spatial and temporal modules. 
For temporal modeling, the advanced multi-scale TCN~\cite{chen2021channel,liu2020disentangling} is used to replace the naive implementation, 
which is capable of modeling actions with multiple durations. 
Despite the improved capacity, the temporal module performs joint-level motion modeling as before, which we find insufficient. 
For spatial modeling, a series of works propose to learn a data-driven refinement for the prescribed graphical structure. 
The refinement can be either channel-agnostic~\cite{shi2019two,zhang2020semantics} (obtained with self-attention mechanism) or channel-specific~\cite{chen2021channel,cheng2020decoupling}. 
Meanwhile, Liu\etal~\cite{liu2020disentangling} introduces multi-scale graph topologies for joint relationship modeling with different ranges. 
There also exist spatial modeling modules that do not require a pre-defined topology. 
Shift-GCN~\cite{cheng2020skeleton} adopts graph shift for inter-joint feature fusion, 
while SGN~\cite{zhang2020semantics} estimates the coefficient matrices based on the input skeleton sequence 
with a lightweight attention module. 
However, the purely dynamic approaches are less competitive and can not achieve the state-of-the-art.
In this work, we design the dynamic group GCN that learns the spatial feature fusion strategy from scratch 
and does not rely on a prescribed graphical structure.
Our DG-GCN outperforms various alternatives in the ablation study.

Besides GCNs, another stream of work adopts convolutional neural networks (CNN) for skeleton-based action recognition. 
Approaches based on 2D-CNN convert a skeleton sequence to a pseudo image and process it with 2D CNN.
This can be obtained by: 1) directly generating a 2D input of shape $V\times T$~\cite{caetano2019skelemotion,ke2017new,li2018co} 
($V$: number of joints; $T$: temporal length) given joint coordinates; 
2) generating pseudo heatmaps for joints (2D only) and aggregating them along the temporal dimension with color coding~\cite{choutas2018potion} or learnable modules~\cite{yan2019pa3d} to form a $K$-channel pseudo image. 
These approaches either fail to exploit the locality nature of CNN or suffer from the information loss during heatmap aggregation,
thus performing worse than representative GCN-based approaches. 
Recently, PoseC3D~\cite{duan2021revisiting} proposes to stack heatmaps along the temporal dimension and use 3D-CNN for skeleton processing. 
Though demonstrating impressive performance, PoseC3D consumes larger amounts of computations and can not be directly applied to 3D skeletons losslessly.
In experiments, we find that while being much lighter, 
our DG-STGCN can achieve better recognition performance than PoseC3D 
when both take high-quality 2D estimated skeletons as inputs.
\section{DG-STGCN}

The enhancement DG-STGCN made lies in two aspects:
1). We design the dynamic group GCN (\textbf{DG-GCN}) and the dynamic group temporal ConvNet (\textbf{DG-TCN}), 
which promote highly flexible and data-driven spatial-temporal modeling of skeleton data.
2). We develop a strong temporal data augmentation strategy, which works universally with different backbones and 
is particularly beneficial to the highly dynamic DG-STGCN.
We first briefly review the pipeline of ST-GCN~\cite{yan2018spatial}, whose architectural design is widely followed by subsequent works.
Then we discuss different designs for both dynamic inter-joint spatial modeling (DG-GCN) and dynamic multi-level temporal modeling (DG-TCN).
Finally, we present the temporal augmentation we adopted and explain its intuition.  

\begin{figure}[t]
    \centering
    \captionsetup{font=footnotesize}
	\includegraphics[width=\linewidth]{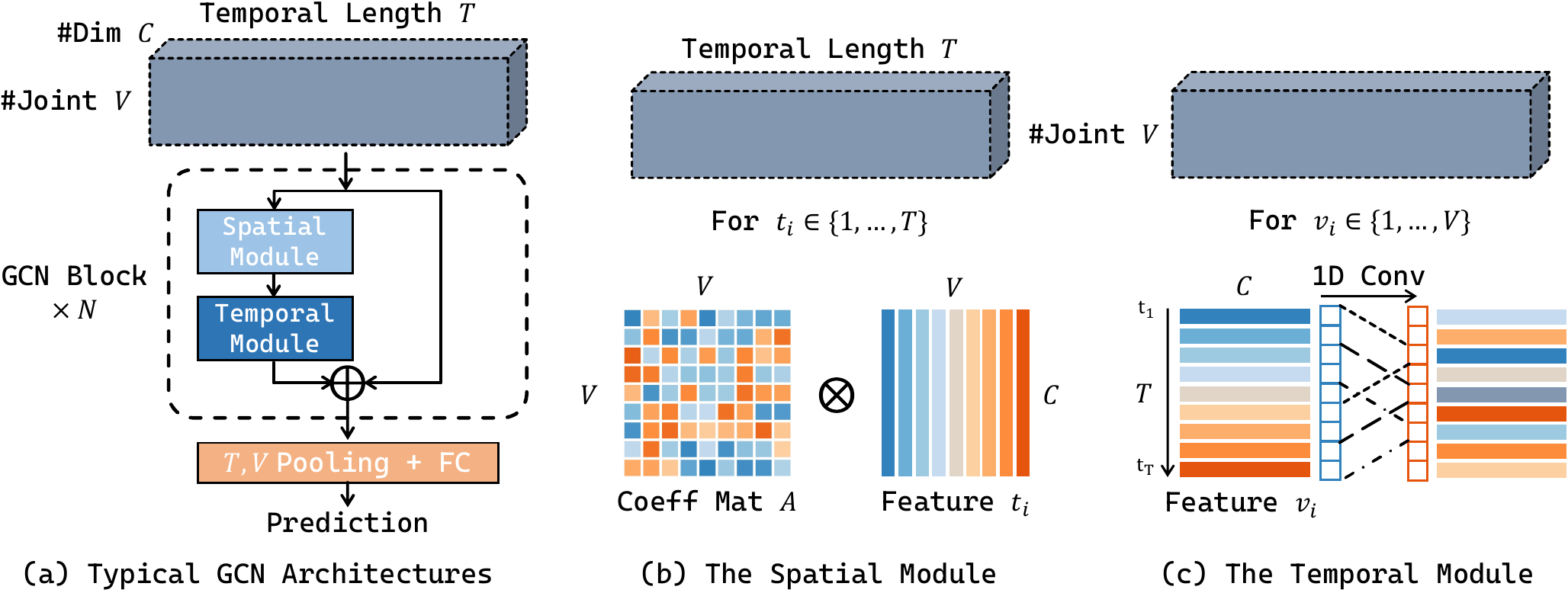} 
	\vspace{-5mm}
	\caption{\textbf{The typical framework of GCNs for skeleton-based action recognition. } 
    (a) A GCN consists of $N$ stacked GCN blocks, each consists of a spatial module and a temporal module. 
    (b) The spatial module performs feature fusion across joints with coefficient matrices $A$ (pre-defined / learned). 
    (c) The temporal module learns temporal features with 1D temporal convolutions. }
	\label{fig-gcnframework}
    \vspace{-3mm}
\end{figure}

\subsection{ST-GCN Recap}

\textbf{Notations. } We use the following notations across this paper.
For inputs and intermediate features $\mX \in \mR^{T\times V\times C}$ of a GCN for skeleton processing, 
$T$ denotes the temporal length; $V$ denotes the number of skeleton joints; $C$ denotes the number of channels (if $\mX$ is the input, $C$ is usually 2 or 3). 
For coefficient matrices $\mA \in \mR^{K\times V\times V}$ used for spatial modeling, 
$K$ denotes the number of matrices (we also call it the number of groups). 
We use $\cF$ to denote the GCN block; $\cG, \cT$ to indicate the spatial module and temporal module, respectively.

In skeleton-based action recognition, 
GCN approaches take a sequence of human joint coordinates (a $T\times V\times C$ tensor $\mX$ for each person) as input and predict the action category. 
Figure~\ref{fig-gcnframework} provides a quick recap of ST-GCN, an early and representative work for skeleton-based action recognition. 
ST-GCN utilizes stacked $N$ GCN blocks ($N=10$) to process the coordinate tensors.  
A GCN block $\cF$ consists of two components: a spatial module $\cG$ and a temporal module $\cT$.  
Given a skeleton feature $\mX \in \mR^{T\times V\times C}$, 
$\cG$ first performs channel inflation and reshaping to get a 4D tensor $\mX' \in \mR^{T\times K\times V\times C}$,
and then uses a set of coefficient matrices $\mA$ for inter-joint spatial modeling: 
\begin{equation}
\cG(\mX)_{lij}  = \sum_{k=1}^{k=K} \sum_{v=1}^{v=V} \mA_{kvi} \mX'_{lkvj}, \qquad\quad \cG(\mX) \in \mR^{T\times V\times C}. 
\end{equation}
The output of $\cG$ is further processed by $\cT$ to learn temporal features. 
The computations of $\cF$ can be summarized as:
\begin{equation}
\cF(\mX) = \cT(\cG(\mX, \mA)) + \mX. 
\end{equation}
Specifically, ST-GCN~\cite{yan2018spatial} adopts sparse matrices derived from a prescribed graphical structure (weighted by learnable importance scores) as $\mA$, 
and instantiates $\cT$ with a single 1D temporal convolution (kernel size 9).
The design of ST-GCN is straightforward and practical, 
widely used in the following works~\cite{chen2021channel,cheng2020skeleton,shi2019skeleton,shi2019two}. 
In this work, we keep the overall architecture unchanged and propose to enhance its capability with dynamic group GCN and dynamic group TCN.  

\subsection{DG-GCN: Dynamic Spatial Modeling from Scratch}
The spatial module $\cG$ fuses features of different joints with coefficient matrices $\mA$, 
while the final recognition performance is highly dependent on the choice of $\mA$. 
One common practice in previous works is to first set $\mA$ as a series of manually defined sparse matrices (derived from the adjacency matrix of joints)
and learn a refinement $\mDelta\mA$ during model training~\cite{chen2021channel,shi2019two,shi2020skeleton}. 
$\mDelta\mA$ can be either static, as a trainable parameter, or dynamic, generated by the model depending on the input sample. 
Though intuitive, we argue that this paradigm is limited in flexibility and may lead to inferior recognition performance than purely data-driven approaches. 

Thus we make the coefficient matrices $\mA$ fully learnable. 
In DG-GCN, $\mA \in \mR^{K\times V \times V}$ is a learnable parameter rather than a set of manually defined matrices. 
For initialization, instead of resorting to derivations of the adjacency matrix, we directly initialize $\mA$ with normal distribution. 
With random initialization, $K$ can be set to an arbitrary value, enabling the novel multi-group design in DG-GCN.

\begin{figure}[t]
    \centering
    \captionsetup{font=small}
	\includegraphics[width=\linewidth]{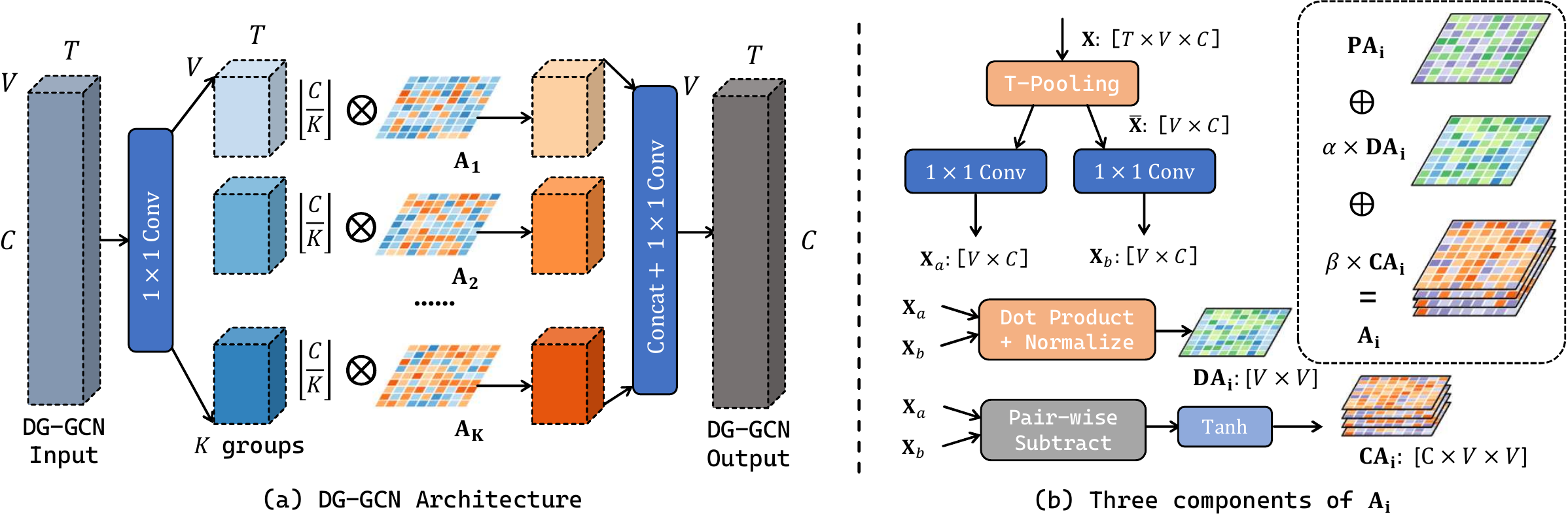} 
	\vspace{-5mm}
	\caption{\textbf{The architecture of the dynamic group GCN (DG-GCN). } }
	\label{fig-dggcn}
    \vspace{-3mm}
\end{figure}

We demonstrate the architecture of DG-GCN in Figure~\ref{fig-dggcn}(a).
In DG-GCN, the coefficient matrices $\mA$ consist of $K$ different components ($K=8$ in experiments).
The input (with $C$ channels) is first processed with a $1\times 1$ convolution, to generate $K$ feature groups (each with $\lfloor\frac{C}{K} \rfloor$ channels). 
We then perform spatial modeling for each feature group independently with the corresponding $\mA_i$. 
Finally, $K$ feature groups are concatenated along the channel dimension and then processed by another $1\times 1$ convolution to generate the output.

Each coefficient matrix $\mA_i$ consists of three learnable components. 
First, a dataset-wise parameter $\mP\mA_i$ (the static term) is learned throughout training. 
Besides, we learn data-depentent refinements $\mD\mA_i$ and $\mC\mA_i$, serving as the dynamic terms. 
The two terms are different in that $\mD\mA_i$ is channel-agnostic while $\mC\mA_i$ is channel-specific.  
Figure~\ref{fig-dggcn}(b) demonstrates how we obtain $\mD\mA_i$ and $\mC\mA_i$. 
Given the input feature $\mX$, temporal pooling is first used to eliminate the dimension $T$.
Two separate $1\times 1$ convolutions are then applied to $\bar{\mX}$ to get $\mX_a$ and $\mX_b$, respectively. 
We define $\mD\mA_i$ as
\begin{equation}
\mD\mA_i = \text{Softmax}(\mX_a \mX_b^{T}, \text{dim=0}),
\end{equation}
and obtain $\mC\mA_i \in \mR^{C\times V\times V}$ by applying the Tanh activation to the pairwise difference between $\mX_a$ and $\mX_b$. 
A weighted sum of three components is used as the final coefficient matrix: 
\begin{equation}
\mA_i = \mP\mA_i + \alpha\times \mD\mA_i + \beta\times \mC\mA_i, 
\end{equation}
$\alpha$, $\beta$ are two learnable parameters. 
In experiments, we find the term $\mP\mA_i$ plays the major role and cannot be omitted, 
while two dynamic terms also steadily contribute to the final recognition performance. 
Thanks to its flexibility and capability, DG-GCN outperforms multiple alternatives in ablation study with a comparable or smaller parameter size and FLOPs. 

\vspace{1mm}
\subsection{DG-TCN: Multi-group TCN with Dynamic Joint-Skeleton Fusion}
\begin{figure}[t]
    \centering
    \begin{minipage}{.58\textwidth}
        \centering
        \captionsetup{font=small}
        \includegraphics[width=\linewidth]{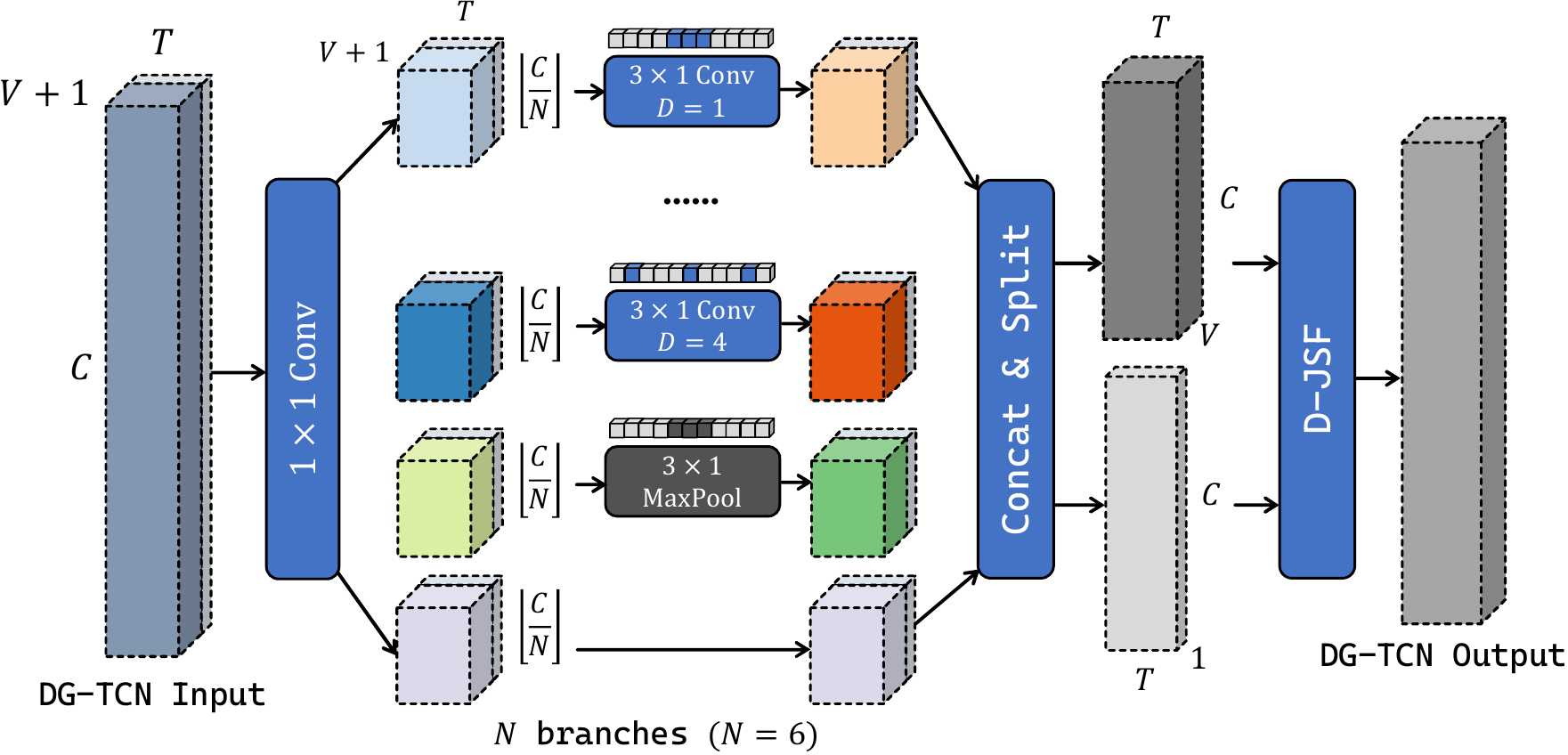} 
        \vspace{-5mm}
        \caption{\textbf{The architecture of the dynamic group TCN (DG-TCN). }  `D' indicates dilation. }
        \label{fig-mstcn}
    \end{minipage}
    \hfill
    \begin{minipage}{.38\textwidth}
        \centering
        \captionsetup{font=small}
        \includegraphics[width=\linewidth]{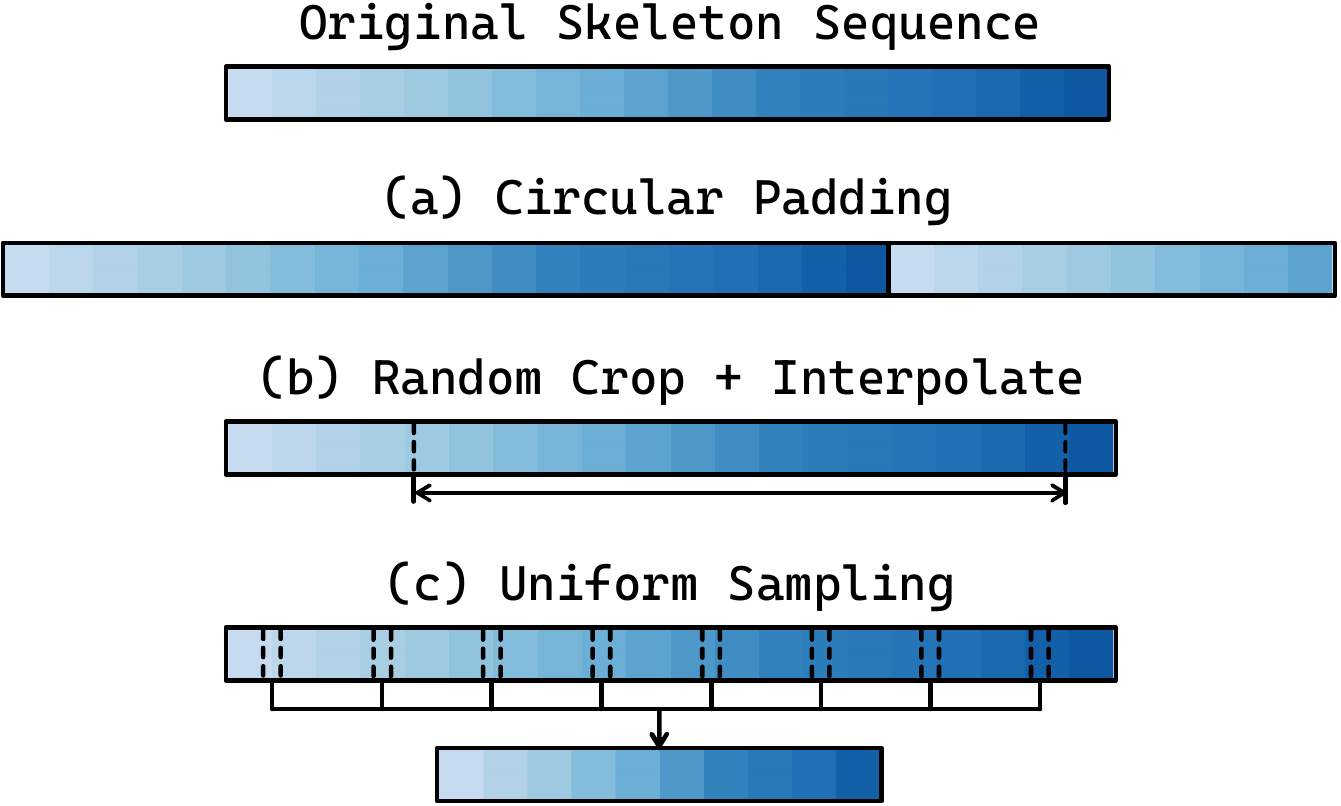} 
        \caption{\textbf{The visualization of Uniform Sampling and two alternatives. } }
        \label{fig-sampling}
    \end{minipage}
    \vspace{-3mm}
\end{figure}

We also adopt the dynamic group-wise design in temporal modeling.   
Following \cite{chen2021channel,liu2020disentangling}, we first replace the vanilla temporal module with a multi-group TCN.
As shown in Figure~\ref{fig-mstcn}, a multi-group TCN consists of multiple branches with different receptive fields.
Each branch performs temporal modeling for a feature group independently. 
Besides saving massive amounts of computation, the temporal modeling capability is also drastically improved with the multi-group design, compared to the vanilla implementation of $\cT$ in ST-GCN. 

We further propose to perform temporal modeling at both joint-level and skeleton-level parallelly. 
Thus we develop the D-JSF module for explicit modeling and dynamically fusing the joint-level and skeleton-level features. 
In D-JSF, we first perform average pooling to $V$ joint-level features $\{\mX_1, ..., \mX_V | \mX_i \in \mR^{C\times T}\}$ to obtain the skeleton-level feature $\mS$.
A multi-group TCN processes both the skeleton feature $\mS$ and $V$ joint features $\mX_i$ in parallel to get $\mS', \mX_i'$. 
Dynamic joint-skeleton fusion is then applied to merge $\mS'$ into each $\mX_i'$. 
Specifically, each DG-TCN instance contains a learned parameter $\gamma \in \mR^V$. 
After the adaptive joint-skeleton fusion, the feature for joint $i$ is $\mX_i' + \gamma_i \mS'$, which will be further processed with a $1\times 1$ convolution.  
In practice, we find that the dynamic fusion with a learnable weight $\gamma$ is the key to the success of DG-TCN.

\subsection{Uniform Sampling as Temporal Data Augmentation}

Thanks to the extensively used dynamic group modules in DG-STGCN, the model capacity is largely improved. 
Meanwhile, the improved flexibility also raises the risk of overfitting, making strong augmentation extremely essential. 

While important, the data augmentation strategies for skeleton data are rarely explored.
In most previous works~\cite{liu2020disentangling,shi2019two,yan2018spatial}, the skeleton sequence is directly padded to a maximum length,
while the padding is either zero or the sequence itself. 
A recent work~\cite{chen2021channel} adopts random cropping as the strategy for temporal augmentation:
crop a substring from the original sequence and resize it to a target length with bilinear interpolation. 
Inspired by \cite{duan2021revisiting}, we first propose to utilize uniform sampling as temporal augmentation. 
To get an $N$-frame sample from a length-$T$ sequence, we first split the sequence into $N$ non-overlapping substrings with equal length. 
Then, we randomly choose one frame from each substring and join them into a new subsequence. 
Uniform sampling can generate much more diversified samples than random cropping, while preserving the integrity of each sample. 
As a strong augmentation, Uniform Sampling consistently outperforms naive padding or random cropping across multiple recognition models and benchmarks. 
The improvement is especially large for DG-STGCN. 
\section{Experiment}

\subsection{Datasets}

To demonstrate the capability and generality of DG-STGCN, 
we conduct extensive experiments on four datasets: NTURGB+D~\cite{liu2020ntu,shahroudy2016ntu}, Kinetics-400~\cite{carreira2017quo}, 
BABEL~\cite{punnakkal2021babel}, and Toyota SmartHome~\cite{Das_2019_ICCV}. 
Following the convention, we report the \textbf{Top-1} accuracy on NTURGB+D and Kinetics-400; report the \textbf{Mean Top-1} accuracy for BABEL and Toyota SmartHome.

\noindent
\textbf{NTURGB+D~\cite{liu2020ntu,shahroudy2016ntu}. } 
NTURGB+D is a large-scale human action recognition dataset collected in indoor environments. 
It has two versions, NTU-60 and NTU-120 (a superset of NTU-60).
NTU-60 contains 57K videos of 60 human actions, while NTU-120 contains 114K videos of 120 human actions. 
The datasets are split in three ways: Cross-subject (X-Sub), Cross-view (X-View, for NTU-60), Cross-setup (X-Set, for NTU-120), for which action subjects, camera views, camera setups are different in training and validation.
For skeleton action recognition, we take the 3D joints as inputs, while our preprocessing follows CTR-GCN~\cite{chen2021channel}. 
To show the generality of DG-STGCN, we also list the performance of models taking 2D joints (estimated with HRNet~\cite{sun2019deep}, provided by \cite{duan2021revisiting}) as inputs.   

\noindent
\textbf{Kinetics-400~\cite{carreira2017quo}. }
Kinetics-400 is a large-scale action recognition dataset with 306K videos and 400 actions. 
For Kinetics-400 experiments, we adopt the 2D joints estimated by OpenPose~\cite{8765346} (provided by ~\cite{yan2018spatial}) as inputs.

\noindent
\textbf{BABEL~\cite{punnakkal2021babel}. }
BABEL is a large-scale dataset for human movement with semantics. 
It annotates $\sim$43.5 hours of mocap from AMASS~\cite{mahmood2019amass}. 
BABEL predicts the 25-joint skeleton used in NTURGB+D from the vertices of the SMPL-H~\cite{romero2022embodied} mesh and uses it for skeleton-based action recognition.
BABEL has two versions, which consist of 60 (BABEL60) and 120 (BABEL120) action categories respectively. 
We report the results on both of them. 

\noindent
\textbf{Toyota SmartHome~\cite{Das_2019_ICCV}. }
Toyota SmartHome is a video dataset focusing on the real-world activities of humans.
The dataset contains 16K videos and 31 different actions in daily life. 
The videos are taken from 7 different camera views.
The dataset is split in two ways: Cross-subject (CS) or Cross-view (CV1: one view for training; CV2: five views for training). 
We only use 3D joints in its Pose V1.2 annotations for training and report results for CS and CV1 splits.

\subsection{Implementation Details} 
DG-STGCN follows the basic settings of ST-GCN. 
Specifically, there are 10 GCN blocks in DG-STGCN, while each GCN block consists of a DG-GCN and a DG-TCN. 
The base channel width is set to 64. 
At the $5_{th}, 8_{th}$ GCN block, we perform temporal pooling to halve the temporal length and also get the channel width doubled.
By default, all models are trained with SGD with momentum 0.9, weight decay 5e$^{-4}$. 
We set the batch size to 128, set the initial learning rate to 0.1, and train the models for 100 epochs with CosineAnnealing learning rate scheduler. 
We use Uniform Sampling to sample subsequences from skeleton data to form training samples.
In the ablation study, we set the input length to 64. 
For comparisons with state-of-the-arts, we set the input length to 100.

\vspace{-3mm}
\subsection{Ablation Study} 

In the ablation study, we train DG-STGCN on two large-scale benchmarks: NTU120-XSub and NTU120-XSet.
We take the joint modality as inputs and follow the preprocessing in \cite{chen2021channel}. 
Our baseline uses GCN with randomly initialized coefficient matrices (3 components) for spatial modeling and a multi-branch TCN (with six branches: 4 kernel-3 1D convolution branches with dilation [1, 2, 3, 4], a kernel-3 max-pooling branch, and a $1\times 1$ convolution branch) for temporal modeling.

\begin{wraptable}{r}{.47\linewidth}
    \captionsetup{font=small, position=top}
	\centering 
    \vspace{-8mm}
    \caption{The topology learned from scratch works well.}
    \vspace{-5mm}
    \label{tab-prelim}
    \resizebox{\linewidth}{!}{
	\tablestyle{4pt}{1.2}
    \begin{tabular}{ccc}
    \\ \shline
    Setting & XSub & XSet \\\shline
    Pre-defined + fix & 83.1 & 84.6 \\ 
    Pre-defined + refine & 83.8 & 85.4 \\ 
    From scratch & 83.6 & 85.3 \\ \shline
    \end{tabular}}
    \vspace{-9mm}
\end{wraptable} 

\vspace{-3mm}
\subsubsection{Preliminary: Is a pre-defined topology indispensable? }
We first conduct pilot experiments to find out how important a prescribed graphical structure is for spatial modeling. 
We consider three alternatives: 1). a fixed topology~\cite{shi2019two} which is pre-defined; 2). \textbf{1} with a static learnable refinement; 
3). a topology learned from scratch, randomly initialized at the beginning.
Table~\ref{tab-prelim} shows that a learnable refinement largely improves the recognition performance. 
However, removing the pre-defined topology only leads to a moderate drop ($\le$ 0.2\%). 
Following Occam's Razor, we do not use any prescribed graphical structure in the following experiments.
This opens a great design space for spatial GCNs.

\vspace{-3mm}
\subsubsection{Dynamic Group GCN (DG-GCN). } 
Thanks to the from-scratch learning of skeleton graphical structures, it is now possible to use an arbitrary number of coefficient matrices for spatial modeling. 
To promote the flexibility of the GCN module, we propose DG-GCN, which divides skeleton features into multiple groups and uses a specific coefficient matrix $\mA_i$ for spatial modeling in each group. 
This section presents the ablation study on the number of groups and the different choices of components in $\mA_i$.  

\begin{table*}[t]
	\centering
	\captionsetup{font=small}
	\captionsetup[subfloat]{font=small, position=bottom}
	% \captionsetup[subffloat]{justification=centering}
	\caption{\small \textbf{Ablation study on different choices for DG-GCN design.}}
    \vspace{1mm}
	\label{tab-dggcn}
	\resizebox{\linewidth}{!}{
	\subfloat[\textbf{Ablation on the number of groups.} \label{tab-dggcn-groups}]{
		\tablestyle{6pt}{1.3}
		\begin{tabular}{ccc}
            \multicolumn{3}{c}{} \\ \shline
			$K$ & XSub & XSet \\ \shline
			4 & 83.7 & 85.2 \\ 
			8 & \textbf{84.1} & \textbf{86.0} \\ 
			12 & 83.9 & 86.0 \\ \shline
		\end{tabular}} 
	\hspace{2mm}
	\subfloat[\textbf{Ablation on each individual component in $\mA$. } 
		\label{tab-dggcn-individual}]{
		\tablestyle{8pt}{1.3}
			\begin{tabular}{ccc} 
                \multicolumn{3}{c}{} \\
				\shline
				Comp. & XSub & XSet\\ \shline
				$\mP\mA$ & \textbf{84.1} & \textbf{86.0} \\
				$\mD\mA$ & 82.3 & 84.8 \\
				$\mC\mA$ & 83.4 & 85.0 \\  \shline
			\end{tabular}}
	\hspace{2mm}
	\subfloat[\textbf{Ablation on different combinations of components in $\mA$.}  
	\label{tab-dggcn-combination}]{
		\tablestyle{4pt}{1.3}
			\begin{tabular}{ccc} 
				\shline
				Comps. & XSub & XSet \\ \shline
				$\mP\mA$ & 84.1 & 86.0 \\
				$\mP\mA + \mD\mA$ & 84.2 & 86.2 \\
				$\mP\mA + \mC\mA$ & 84.8 & 86.7 \\  
                $\mP\mA + \mD\mA + \mC\mA$ & \textbf{85.1} & \textbf{87.2} \\ \shline
			\end{tabular}}}
    \vspace{-13mm}
\end{table*}

\noindent
\textbf{Ablation on the group number. }
We first explore using different values for the hyper-parameter $K$, which is the number of groups in DG-GCN.
Since the channel width of each group is proportional to $1/K$, adjusting the value of $K$ does not affect either the parameter size or the computation amounts consumed.
We gradually increase $K$ from 4 to 12 and list the recognition performance in Table~\ref{tab-dggcn-groups}. 
More groups improve the flexibility of DG-GCN and benefit the skeleton spatial modeling. 
Empirically, we find that eight groups lead to a reasonably good result (adding more groups does not help a lot).
Thus we use eight groups in the following experiments. 

\noindent
\textbf{Ablation on the topology components. }
As a network parameter, the learned from-scratch topology is a static component ($\mP\mA$) in the final coefficient matrices. 
In DG-GCN, we also incorporate two dynamic terms, namely $\mD\mA$ (channel-agnostic) and $\mC\mA$ (channel-specific). 
Two dynamic components share the same encoding layers (two $1\times 1$ convolutions), 
which means all models with dynamic components share a similar parameter size (1.69M) and FLOPs (1.65G). 
Compared to using only the static component (1.25M params, 1.63G FLOPs), 
using dynamic terms increases the parameter size by $1/3$, while the computation consumed almost remains the same. 
We first evaluate each component individually and show the recognition performance in Table~\ref{tab-dggcn-individual}. 
Dynamic terms used alone cause severe degradation in recognition performance compared to the static term $\mP\mA$, 
which partially explains why previous purely dynamic approaches (like SGN~\cite{zhang2020semantics}) can not beat the state-of-the-art. 
We further investigate the effects of different combinations of terms and present the results in Table~\ref{tab-dggcn-combination}. 
For spatial modeling, the channel-specific term $\mC\mA$ plays a more critical role than the channel-agnostic term $\mD\mA$. 
Combining three components leads to the best recognition results.

\noindent
\textbf{How much data does DG-GCN require? }  
Learning joint relationships solely from data without any prior knowledge, 
we wonder how much data DG-GCN requires to perform well. 
Thus we conduct a series of experiments on the NTU120-XSub split, changing the amounts of training data to see what happens. 
The data amount we use covers a wide range: from $1/16$ of the original training set to the full scale. 
Figure~\ref{fig-dataamount} compares the three spatial modeling strategies with different amounts of training data.  
A pre-defined fixed topology performs worst across all experiments. 
When the training set is of small scale (\emph{i.e.,} smaller than $1/4$ of NTU120-XSub training set), 
a manually defined topology is beneficial to the spatial modeling of skeleton data.
However, as the training set grows, the advantage of DG-GCN gradually gets larger. 
Experiment results demonstrate that the potential of DG-GCN is promising if trained on large-scale data. 

\noindent
\begin{table*}[t]
    \centering
	\captionsetup{font=small}
    \begin{minipage}[b]{.53\linewidth}
        \centering
        \includegraphics[width=.93\linewidth]{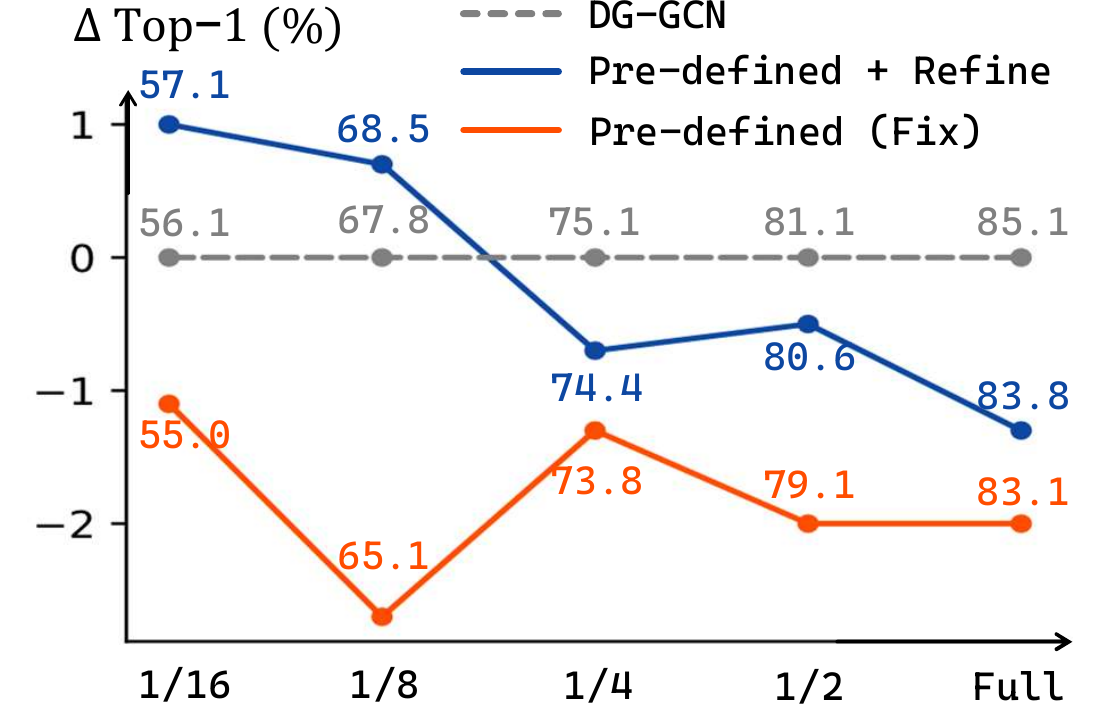}
        \vspace{1mm}
		\captionof{figure}{\textbf{Spatial modeling strategies with different amounts of training data. }
		DG-GCN's advantage grows with the data scale. }
		\label{fig-dataamount}
	\end{minipage}
    \hfill
	\begin{minipage}[b]{.44\linewidth}
		\centering
        \resizebox{\linewidth}{!}{
	        \tablestyle{4pt}{1.3}
            \begin{tabular}{cccc}
            \shline
             & Xsub & XSet & GFLOPs\\ \shline
            TCN (K9) & 83.1 & 84.7 & 3.46 \\
            Multi-Group TCN & 83.6 & 85.3 & 1.63 \\ 
            w. JSF (concat) & 83.5 & 86.0 & 1.94 \\ 
            w. JSF (sum) & 83.2 & 85.7 & 1.65\\ 
            w. D-JSF & \textbf{84.0} & \textbf{86.2} & 1.65\\ \shline
            \end{tabular}}
        \vspace{1mm}
		\captionof{table}{\textbf{Ablation on dynamic temporal modeling. } 
            The Multi-Group TCN improves temporal modeling capability with reduced computations, 
            while D-JSF promotes dynamic joint-skeleton fusion with minimal additional cost. }
		\label{tab-temporal}
	\end{minipage}
    \vspace{-8mm}
\end{table*}

\vspace{-10mm}
\subsubsection{Dynamic Group TCN (DG-TCN). }
We conduct ablation experiments to validate our design choice for the temporal module. 
Instead of the vanilla implementation (a single 1D convolution with kernel 9) in ST-GCN, we adopt the DG-TCN for temporal modeling.
DG-TCN consists of multiple branches with different receptive fields for dynamic temporal modeling.
It is also equipped with the dynamic joint-skeleton fusion module (D-JSF) for adaptive fusion between joint motion and skeleton motion. 
Table~\ref{tab-temporal} shows that, besides improving the temporal modeling capability, the multi-group design also substantially reduces the computation cost by using a small channel width for every single branch\footnote{The parameter size is also significantly reduced: 2.99M $\rightarrow$ 1.25M}. 
For the JSF module, besides fusing the skeleton-level feature with joint-level feature dynamically with learnable coefficients, 
we also test two simpler alternatives, \emph{e.g.}, directly concatenating or summing up the two features. 
From Table~\ref{tab-temporal}, we see that D-JSF performs the best among the three alternatives. 
It's worth noting that D-JSF is an efficient module that only consumes 1\% additional computation cost of the backbone model. 

\begin{table}[t]
    \captionsetup{font=small, position=top}
	\centering 
    \caption{\textbf{Validating Uniform Sampling and two alternatives adopted in previous works on multiple backbones. 
    $\Delta$ denotes the gain from Uniform Sampling compared to the baseline without temporal augmentation. }}
    \vspace{1mm}
	\label{tab-aug}
	\resizebox{\linewidth}{!}{
	\tablestyle{4pt}{1.4}
    \begin{tabular}{c|cccc|cccc}
    \shline
    & \multicolumn{4}{c|}{NTU120-XSub} & \multicolumn{4}{c}{NTU120-XSet} \\
    Aug & None & Random Crop & Uni-Sample & $\Delta$ & None & Random Crop & Uni-Sample & $\Delta$ \\ \shline
    ST-GCN~\cite{yan2018spatial} & 81.1 & 82.2 & 82.2 & 1.1 & 84.8 & 83.9 & 85.8 & 1.0 \\
    AGCN~\cite{shi2019two} & 81.6 & 82.3 & 82.8 & 1.2 & 84.0 & 84.8 & 85.3 & 1.3 \\
    MS-G3D~\cite{liu2020disentangling} & 83.3 & 83.1 & 84.2 & 0.9 & \textbf{85.7} & 85.0 & 85.8 & 0.1 \\
    CTR-GCN~\cite{chen2021channel} & 82.4 & 83.4 & 83.8 & 1.4 & 85.2 & 85.6 & 85.8 & 0.6 \\
    DG-STGCN & \textbf{83.5} & \textbf{84.0} & \textbf{85.3} & \textbf{1.8} & 85.1 & \textbf{86.5} & \textbf{87.5} & \textbf{2.4} \\ \shline
    \end{tabular}}
    \vspace{-6mm}
\end{table}

\vspace{-5mm}
\subsubsection{Uniform Sampling as Temporal Data Augmentation. }
With the improved network capacity and flexibility, DG-STGCN requires stronger data augmentations to work well.
Inspired by \cite{duan2021revisiting}, we propose to use Uniform Sampling as a temporal data augmentation strategy, which samples subsequences from skeleton data to form training samples. 
In Table~\ref{tab-aug}, we compare Uniform Sampling with two alternative practices adopted in previous works: 
1). No temporal augmentation~\cite{liu2020disentangling,shi2019two}: circularly padding a skeleton sequence to a maximum length (300 for NTU); 
2). Random Cropping~\cite{chen2021channel}: cropping a substring from the skeleton sequence and interpolating it to the given length 64. 
We find that Uniform Sampling is a \textbf{general} data augmentation strategy, leading to steady improvement across all backbones and benchmarks we tested. 
Specifically, the dynamic DG-STGCN obtains the most significant gain with the strong data augmentation.

\vspace{-3mm}
\subsection{Comparisons with the State-of-the-Art. }
For comparisons with state-of-the-art approaches, we adopt a longer training schedule. 
We train each model with the CosineAnnealing scheduler for 150 epochs. 
We use Uniform Sampling to form each training sample and set the input length to 100 by default.
For Toyota SmartHome, which consists of long sequences up to thousands of frames, we set the input length to 200.
We report the Top-1 accuracy for NTURGB+D and Kinetics-400 and the mean per-class accuracy for BABEL and Toyota SmartHome (two highly unbalanced datasets).

\begin{table}[t]
    \captionsetup{font=small, position=top}
	\centering 
    \caption{\textbf{Classification accuracy comparisons with state-of-the-art methods on NTURGB+D and Kinetics-400. The * notation means using 2D pose estimation results provided by \cite{duan2021revisiting}. }}
    \vspace{1mm}
	\label{tab-sota}
	\resizebox{\linewidth}{!}{
	\tablestyle{6pt}{1.2}
    \begin{tabular}{c|ccccc}
    \shline
    & NTU60-XSub & NTU60-XView & NTU120-XSub & NTU120-XSet & Kinetics \\ \shline
    ST-GCN~\cite{yan2018spatial}    & 81.5  & 88.3  & 70.7  & 73.2  & 30.7 \\
    SGN~\cite{zhang2020semantics}   & 86.6  & 93.4  & -     & -     & -    \\
    AS-GCN~\cite{li2019actional}    & 86.8  & 94.2  & 78.3  & 79.8  & 34.8 \\
    RA-GCN~\cite{song2020richly}    & 87.3  & 93.6  & 78.3  & 79.8  & 34.8 \\
    AGCN~\cite{shi2019two}  & 88.5  & 95.1  & -     & -     & 36.1 \\
    DGNN~\cite{shi2019skeleton} & 89.9  & 96.1  & -     & -     & 36.9 \\
    FGCN~\cite{yang2021feedback}    & 90.2  & 96.3  & 85.4  & 87.4  & - \\
    ShiftGCN~\cite{cheng2020skeleton}   & 90.7  & 96.5  & 85.9  & 87.6  & - \\
    DSTA-Net~\cite{shi2020decoupled}    & 91.5  & 96.4  & 86.6  & 89.0  & - \\
    MS-G3D~\cite{liu2020disentangling}  & 91.5  & 96.2  & 86.9  & 88.4  & 38.0 \\
    CTR-GCN~\cite{chen2021channel}  & 92.4  & 96.8  & 88.9  & 90.6  & - \\ 
    ST-GCN++~\cite{duan2022pyskl} & 92.6 & 97.4 & 88.6 & 90.8 & \\ \shline
    2s DG-STGCN & 92.9  & 97.3  & 89.2  & 91.2  & 39.5 \\
    DG-STGCN & \textbf{93.2}  & \textbf{97.5}  & \textbf{89.6}  & \textbf{91.4}  & \textbf{40.3} \\ \shline
    & NTU60-XSub* & NTU60-XView* & NTU120-XSub* & NTU120-XSet* &  \\ \shline
    MS-G3D++~\cite{liu2020disentangling} & 92.2 & 96.6 & 87.2 & 89.0 & \\ 
    PoseC3D~\cite{duan2021revisiting} & 94.1 & 97.1 & 86.9 & 90.3 & \\
    ST-GCN++~\cite{duan2022pyskl} & 93.2 & 98.5 & 86.4 & 90.3 & \\ \shline
    2s DG-STGCN & 93.9 & 98.5 & 87.3 & 91.1 & \\
    DG-STGCN & \textbf{94.1} & \textbf{98.6} & \textbf{87.5} & \textbf{91.3} & \\ \shline
    \end{tabular}}
    \vspace{-5mm}
\end{table}

\vspace{-3mm}
\subsubsection{Results on NTURGB+D and Kinetics-400. }

For NTURGB+D and Kinetics-400 experiments, we follow the widely used practice~\cite{chen2021channel,shi2020skeleton} to train models 
on four skeleton modalities (joint, bone, joint motion, and bone motion) and report the recognition performance of the ensemble for a fair comparison.
We compare our models with the state-of-the-art methods on NTURGB+D and Kinetics-400 in Table~\ref{tab-sota} (upper).
Our DG-STGCN outperforms \textbf{all} existing methods across four NTURGB+D benchmarks. 
Specifically, with joint\texttt{+}bone fusion (2s) only, 
our approach outperforms the current state-of-the-art CTR-GCN using four modality fusion with a notable margin. 
Fusing predictions of four skeleton modalities, we significantly improve the accuracy on four NTURGB+D benchmarks by nearly 1\%. 
On Kinetics-400, DG-STGCN surpasses the state-of-the-art MS-G3D by 2.3\% in Top-1 Acc. 
To demonstrate its generality, we also train DG-STGCN with high-quality 2D estimated skeleton~\cite{duan2021revisiting}. 
Results in Table~\ref{tab-sota} (lower) show that while consuming almost $10\times$ less computation, DG-STGCN achieves better performance than the 3D-CNN-based PoseC3D.

\begin{table}[t]
    \captionsetup{font=small, position=top}
	\centering 
    \begin{minipage}{.45\linewidth}
        \caption{\textbf{Results on BABEL.} We report both bone-only and joint\texttt{+}bone fusion (2s) performance for DG-STGCN. }
        \vspace{1mm}
        \centering
        \resizebox{.98\linewidth}{!}{
	        \tablestyle{4pt}{1.4}
            \begin{tabular}{ccc}
            \shline
            BABEL60 & Mean Acc & Top-1 \\ \shline
            2s-AGCN~\cite{shi2019two} & 30.4 & 33.4 \\ 
            DG-STGCN (Bone) & 36.2 & 39.0 \\ 
            2s DG-STGCN & \textbf{38.1} & \textbf{40.0} \\ \shline
            BABEL120 & Mean Acc & Top-1 \\ \shline
            2s-AGCN~\cite{shi2019two} & 26.2 & 27.9 \\
            DG-STGCN (Bone) & 32.3 & 31.3 \\ 
            2s DG-STGCN & \textbf{33.7} & \textbf{32.8} \\ \shline
            \end{tabular}}
    \end{minipage}%
    \hfill
    \begin{minipage}{.53\linewidth}
        \caption{\textbf{Results on Toyota SmartHome.} 
        \texttt{++} indicates using high-quality 2D Pose. 
        * denotes being pre-trained on another dataset.}
        \vspace{1mm}
        \centering
        \resizebox{\linewidth}{!}{
	        \tablestyle{8pt}{1.3}
            \begin{tabular}{ccc}
            \shline
             & CS & CV1 \\ \shline
            LSTM~\cite{mahasseni2016regularizingls} & 42.5 & 13.4 \\ 
            MS-AAGCN~\cite{shi2019skeleton} & 56.5 & - \\
            2s-AGCN\texttt{++} ~\cite{shi2019two} & 57.1 & 22.1 \\
            2s-AGCN\texttt{++} w. PRS~\cite{yang2021selectivesa} & 60.9 & 22.5 \\
            2s-UNIK\texttt{++}* w. PRS~\cite{yang2021unikau} & 64.3 & \textbf{36.1} \\
            2s DG-STGCN & \textbf{64.8} & 23.2 \\ 
            2s DG-STGCN* & \textbf{65.1} & \textbf{41.8} \\ \shline
            \end{tabular}}
    \end{minipage} 
    \vspace{-4mm}
\end{table}

\vspace{-1mm}
\subsubsection{Results on BABEL and Toyota SmartHome. }
We also train DG-STGCN on BABEL and Toyota SmartHome to test its generality.
BABEL~\cite{punnakkal2021babel} predicts the 25 joints defined by NTURGB+D based on meshes of AMASS mocap data. 
Thus there exists a distribution shift for joint coordinates compared to the original NTURGB+D. 
Meanwhile, Toyota SmartHome~\cite{Das_2019_ICCV} contains activities of daily living and provide 13 3D joints estimated with LCRNet\texttt{++}~\cite{rogez2019lcr}. 
Following \cite{punnakkal2021babel}, we train DG-STGCN with a class-balanced focal loss on BABEL60 and BABEL120. 
On both benchmarks, DG-STGCN with single bone modality outperforms the 2s-AGCN baseline by around 6\% in mean per-class accuracy. 
With joint\texttt{+}bone fusion, the gap is extended to $>7$\%.
On Toyota SmartHome \textbf{CS} split, DG-STGCN trained with 3D keypoints outperform all previous state-of-the-arts 
in skeleton-based action recognition (including entries that take high-quality 2D poses as inputs).
With NTU120-XSet pre-training (with 54K training samples), 
DG-STGCN surpasses UNIK pretrained on Posetics~\cite{yang2021unikau} with 142K training samples on the \textbf{CV1} split. 
The results demonstrate the impressive transferring capability of DG-STGCN:
it works even when the definitions of the joint sets are entirely different for the source dataset and the target dataset 
(25 joints in NTURGB+D \emph{vs.} 13 joints in Toyota SmartHome). 

\section{Conclusion}

This work presents a highly dynamic framework DG-STGCN for the spatial-temporal modeling of skeleton data.
DG-STGCN consists of two novel modules with dynamic multi-group design, namely DG-GCN and DG-TCN. 
For spatial modeling, 
DG-GCN fuses joint-level features with coefficient matrices learned from scratch 
and does not rely on prescribed graphical structures. 
Meanwhile, DG-TCN adopts the group-wise temporal convolution with diversified receptive fields and 
the dynamic joint-skeleton feature module for dynamic multi-level temporal modeling. 
DG-STGCN surpasses the state-of-the-art on four challenging benchmarks, 
demonstrating its impressive capability and effectiveness. 

\clearpage

\bibliographystyle{splncs04}
\bibliography{egbib}

\begin{thebibliography}{10}
\providecommand{\url}[1]{\texttt{#1}}
\providecommand{\urlprefix}{URL }
\providecommand{\doi}[1]{https://doi.org/#1}

\bibitem{bruna2013spectral}
Bruna, J., Zaremba, W., Szlam, A., LeCun, Y.: Spectral networks and locally
  connected networks on graphs. arXiv preprint arXiv:1312.6203  (2013)

\bibitem{caetano2019skelemotion}
Caetano, C., Sena, J., Br{\'e}mond, F., Dos~Santos, J.A., Schwartz, W.R.:
  Skelemotion: A new representation of skeleton joint sequences based on motion
  information for 3d action recognition. In: AVSS. pp.~1--8. IEEE (2019)

\bibitem{8765346}
{Cao}, Z., {Hidalgo Martinez}, G., {Simon}, T., {Wei}, S., {Sheikh}, Y.A.:
  Openpose: Realtime multi-person 2d pose estimation using part affinity
  fields. TPAMI  (2019)

\bibitem{carreira2017quo}
Carreira, J., Zisserman, A.: Quo vadis, action recognition? a new model and the
  kinetics dataset. In: CVPR. pp. 6299--6308 (2017)

\bibitem{chen2021channel}
Chen, Y., Zhang, Z., Yuan, C., Li, B., Deng, Y., Hu, W.: Channel-wise topology
  refinement graph convolution for skeleton-based action recognition. In: ICCV.
  pp. 13359--13368 (2021)

\bibitem{cheng2020decoupling}
Cheng, K., Zhang, Y., Cao, C., Shi, L., Cheng, J., Lu, H.: Decoupling gcn with
  dropgraph module for skeleton-based action recognition. In: ECCV. pp.
  536--553. Springer (2020)

\bibitem{cheng2020skeleton}
Cheng, K., Zhang, Y., He, X., Chen, W., Cheng, J., Lu, H.: Skeleton-based
  action recognition with shift graph convolutional network. In: CVPR. pp.
  183--192 (2020)

\bibitem{choutas2018potion}
Choutas, V., Weinzaepfel, P., Revaud, J., Schmid, C.: Potion: Pose motion
  representation for action recognition. In: CVPR. pp. 7024--7033 (2018)

\bibitem{Das_2019_ICCV}
Das, S., Dai, R., Koperski, M., Minciullo, L., Garattoni, L., Bremond, F.,
  Francesca, G.: Toyota smarthome: Real-world activities of daily living. In:
  ICCV (October 2019)

\bibitem{defferrard2016convolutional}
Defferrard, M., Bresson, X., Vandergheynst, P.: Convolutional neural networks
  on graphs with fast localized spectral filtering. NeurIPS  \textbf{29} (2016)

\bibitem{duan2022pyskl}
Duan, H., Wang, J., Chen, K., Lin, D.: Pyskl: Towards good practices for
  skeleton action recognition. arXiv preprint arXiv:2205.09443  (2022)

\bibitem{duan2021revisiting}
Duan, H., Zhao, Y., Chen, K., Shao, D., Lin, D., Dai, B.: Revisiting
  skeleton-based action recognition. arXiv preprint arXiv:2104.13586  (2021)

\bibitem{duan2020omni}
Duan, H., Zhao, Y., Xiong, Y., Liu, W., Lin, D.: Omni-sourced webly-supervised
  learning for video recognition. In: ECCV. pp. 670--688. Springer (2020)

\bibitem{duvenaud2015convolutional}
Duvenaud, D.K., Maclaurin, D., Iparraguirre, J., Bombarell, R., Hirzel, T.,
  Aspuru-Guzik, A., Adams, R.P.: Convolutional networks on graphs for learning
  molecular fingerprints. NeurIPS  \textbf{28} (2015)

\bibitem{feichtenhofer2020x3d}
Feichtenhofer, C.: X3d: Expanding architectures for efficient video
  recognition. In: CVPR. pp. 203--213 (2020)

\bibitem{feichtenhofer2019slowfast}
Feichtenhofer, C., Fan, H., Malik, J., He, K.: Slowfast networks for video
  recognition. In: ICCV. pp. 6202--6211 (2019)

\bibitem{hamilton2017inductive}
Hamilton, W., Ying, Z., Leskovec, J.: Inductive representation learning on
  large graphs. NeurIPS  \textbf{30} (2017)

\bibitem{hammond2011wavelets}
Hammond, D.K., Vandergheynst, P., Gribonval, R.: Wavelets on graphs via
  spectral graph theory. ACHA  \textbf{30}(2),  129--150 (2011)

\bibitem{henaff2015deep}
Henaff, M., Bruna, J., LeCun, Y.: Deep convolutional networks on
  graph-structured data. arXiv preprint arXiv:1506.05163  (2015)

\bibitem{hussein2013human}
Hussein, M.E., Torki, M., Gowayyed, M.A., El-Saban, M.: Human action
  recognition using a temporal hierarchy of covariance descriptors on 3d joint
  locations. In: IJCAI (2013)

\bibitem{jhuangICCV2013}
Jhuang, H., Gall, J., Zuffi, S., Schmid, C., Black, M.J.: Towards understanding
  action recognition. In: ICCV. pp. 3192--3199 (Dec 2013)

\bibitem{ke2017new}
Ke, Q., Bennamoun, M., An, S., Sohel, F., Boussaid, F.: A new representation of
  skeleton sequences for 3d action recognition. In: CVPR. pp. 3288--3297 (2017)

\bibitem{kim2017interpretable}
Kim, T.S., Reiter, A.: Interpretable 3d human action analysis with temporal
  convolutional networks. In: CVPRW. pp. 1623--1631. IEEE (2017)

\bibitem{kipf2016semi}
Kipf, T.N., Welling, M.: Semi-supervised classification with graph
  convolutional networks. arXiv preprint arXiv:1609.02907  (2016)

\bibitem{li2018co}
Li, C., Zhong, Q., Xie, D., Pu, S.: Co-occurrence feature learning from
  skeleton data for action recognition and detection with hierarchical
  aggregation. arXiv preprint arXiv:1804.06055  (2018)

\bibitem{li2019actional}
Li, M., Chen, S., Chen, X., Zhang, Y., Wang, Y., Tian, Q.: Actional-structural
  graph convolutional networks for skeleton-based action recognition. In: CVPR.
  pp. 3595--3603 (2019)

\bibitem{liu2020ntu}
Liu, J., Shahroudy, A., Perez, M., Wang, G., Duan, L.Y., Kot, A.C.: Ntu rgb+d
  120: A large-scale benchmark for 3d human activity understanding. TPAMI
  \textbf{42}(10),  2684--2701 (2020)

\bibitem{liu2016spatio}
Liu, J., Shahroudy, A., Xu, D., Wang, G.: Spatio-temporal lstm with trust gates
  for 3d human action recognition. In: ECCV. pp. 816--833. Springer (2016)

\bibitem{liu2020disentangling}
Liu, Z., Zhang, H., Chen, Z., Wang, Z., Ouyang, W.: Disentangling and unifying
  graph convolutions for skeleton-based action recognition. In: CVPR. pp.
  143--152 (2020)

\bibitem{mahasseni2016regularizingls}
Mahasseni, B., Todorovic, S.: Regularizing long short term memory with 3d
  human-skeleton sequences for action recognition. CVPR pp. 3054--3062 (2016)

\bibitem{mahmood2019amass}
Mahmood, N., Ghorbani, N., Troje, N.F., Pons-Moll, G., Black, M.J.: Amass:
  Archive of motion capture as surface shapes. In: ICCV. pp. 5442--5451 (2019)

\bibitem{piergiovanni2019representation}
Piergiovanni, A., Ryoo, M.S.: Representation flow for action recognition. In:
  CVPR. pp. 9945--9953 (2019)

\bibitem{punnakkal2021babel}
Punnakkal, A.R., Chandrasekaran, A., Athanasiou, N., Quiros-Ramirez, A., Black,
  M.J.: Babel: Bodies, action and behavior with english labels. In: CVPR. pp.
  722--731 (2021)

\bibitem{rogez2019lcr}
Rogez, G., Weinzaepfel, P., Schmid, C.: Lcr-net++: Multi-person 2d and 3d pose
  detection in natural images. TPAMI  \textbf{42}(5),  1146--1161 (2019)

\bibitem{romero2022embodied}
Romero, J., Tzionas, D., Black, M.J.: Embodied hands: Modeling and capturing
  hands and bodies together. arXiv preprint arXiv:2201.02610  (2022)

\bibitem{sevilla2018integration}
Sevilla-Lara, L., Liao, Y., G{\"u}ney, F., Jampani, V., Geiger, A., Black,
  M.J.: On the integration of optical flow and action recognition. In: GCPR.
  pp. 281--297. Springer (2018)

\bibitem{shahroudy2016ntu}
Shahroudy, A., Liu, J., Ng, T.T., Wang, G.: Ntu rgb+d: A large scale dataset
  for 3d human activity analysis. In: CVPR. pp. 1010--1019 (2016)

\bibitem{shi2019skeleton}
Shi, L., Zhang, Y., Cheng, J., Lu, H.: Skeleton-based action recognition with
  directed graph neural networks. In: CVPR. pp. 7912--7921 (2019)

\bibitem{shi2019two}
Shi, L., Zhang, Y., Cheng, J., Lu, H.: Two-stream adaptive graph convolutional
  networks for skeleton-based action recognition. In: CVPR. pp. 12026--12035
  (2019)

\bibitem{shi2020decoupled}
Shi, L., Zhang, Y., Cheng, J., Lu, H.: Decoupled spatial-temporal attention
  network for skeleton-based action recognition. arXiv preprint
  arXiv:2007.03263  (2020)

\bibitem{shi2020skeleton}
Shi, L., Zhang, Y., Cheng, J., Lu, H.: Skeleton-based action recognition with
  multi-stream adaptive graph convolutional networks. TIP  \textbf{29},
  9532--9545 (2020)

\bibitem{simonyan2014two}
Simonyan, K., Zisserman, A.: Two-stream convolutional networks for action
  recognition in videos. NeurIPS  \textbf{27} (2014)

\bibitem{song2017end}
Song, S., Lan, C., Xing, J., Zeng, W., Liu, J.: An end-to-end spatio-temporal
  attention model for human action recognition from skeleton data. In: AAAI.
  vol.~31 (2017)

\bibitem{song2020richly}
Song, Y.F., Zhang, Z., Shan, C., Wang, L.: Richly activated graph convolutional
  network for robust skeleton-based action recognition. TCSVT  \textbf{31}(5),
  1915--1925 (2020)

\bibitem{sun2019deep}
Sun, K., Xiao, B., Liu, D., Wang, J.: Deep high-resolution representation
  learning for human pose estimation. In: CVPR. pp. 5693--5703 (2019)

\bibitem{tran2019video}
Tran, D., Wang, H., Torresani, L., Feiszli, M.: Video classification with
  channel-separated convolutional networks. In: ICCV. pp. 5552--5561 (2019)

\bibitem{tran2018closer}
Tran, D., Wang, H., Torresani, L., Ray, J., LeCun, Y., Paluri, M.: A closer
  look at spatiotemporal convolutions for action recognition. In: CVPR. pp.
  6450--6459 (2018)

\bibitem{velivckovic2017graph}
Veli{\v{c}}kovi{\'c}, P., Cucurull, G., Casanova, A., Romero, A., Lio, P.,
  Bengio, Y.: Graph attention networks. arXiv preprint arXiv:1710.10903  (2017)

\bibitem{vemulapalli2014human}
Vemulapalli, R., Arrate, F., Chellappa, R.: Human action recognition by
  representing 3d skeletons as points in a lie group. In: CVPR. pp. 588--595
  (2014)

\bibitem{wang2012mining}
Wang, J., Liu, Z., Wu, Y., Yuan, J.: Mining actionlet ensemble for action
  recognition with depth cameras. In: CVPR. pp. 1290--1297. IEEE (2012)

\bibitem{xu2018powerful}
Xu, K., Hu, W., Leskovec, J., Jegelka, S.: How powerful are graph neural
  networks? arXiv preprint arXiv:1810.00826  (2018)

\bibitem{yan2019pa3d}
Yan, A., Wang, Y., Li, Z., Qiao, Y.: Pa3d: Pose-action 3d machine for video
  recognition. In: CVPR. pp. 7922--7931 (2019)

\bibitem{yan2018spatial}
Yan, S., Xiong, Y., Lin, D.: Spatial temporal graph convolutional networks for
  skeleton-based action recognition. In: AAAI (2018)

\bibitem{yang2021selectivesa}
Yang, D., Dai, R., Wang, Y., Mallick, R., Minciullo, L., Francesca, G.,
  Br{\'e}mond, F.: Selective spatio-temporal aggregation based pose refinement
  system: Towards understanding human activities in real-world videos. WACV pp.
  2362--2371 (2021)

\bibitem{yang2021unikau}
Yang, D., Wang, Y., Dantcheva, A., Garattoni, L., Francesca, G., Br{\'e}mond,
  F.: Unik: A unified framework for real-world skeleton-based action
  recognition. ArXiv  \textbf{abs/2107.08580} (2021)

\bibitem{yang2021feedback}
Yang, H., Yan, D., Zhang, L., Sun, Y., Li, D., Maybank, S.J.: Feedback graph
  convolutional network for skeleton-based action recognition. TIP
  \textbf{31},  164--175 (2021)

\bibitem{ye2020dynamic}
Ye, F., Pu, S., Zhong, Q., Li, C., Xie, D., Tang, H.: Dynamic gcn:
  Context-enriched topology learning for skeleton-based action recognition. In:
  MM. pp. 55--63 (2020)

\bibitem{zhang2020semantics}
Zhang, P., Lan, C., Zeng, W., Xing, J., Xue, J., Zheng, N.: Semantics-guided
  neural networks for efficient skeleton-based human action recognition. In:
  CVPR. pp. 1112--1121 (2020)

\bibitem{zhang2017geometric}
Zhang, S., Liu, X., Xiao, J.: On geometric features for skeleton-based action
  recognition using multilayer lstm networks. In: WACV. pp. 148--157. IEEE
  (2017)

\bibitem{zhang2012microsoft}
Zhang, Z.: Microsoft kinect sensor and its effect. IEEE multimedia
  \textbf{19}(2),  4--10 (2012)

\bibitem{zhu2016co}
Zhu, W., Lan, C., Xing, J., Zeng, W., Li, Y., Shen, L., Xie, X.: Co-occurrence
  feature learning for skeleton based action recognition using regularized deep
  lstm networks. In: AAAI. vol.~30 (2016)

\end{thebibliography}

\end{document}